\pdfoutput=1

\documentclass[11pt]{article}

\usepackage[final]{acl}
\usepackage{times}
\usepackage{latexsym}
\usepackage[T1]{fontenc}
\usepackage[utf8]{inputenc}
\usepackage{microtype}
\usepackage{inconsolata}
\usepackage{graphicx}
\usepackage{enumitem}
\usepackage{booktabs}
\usepackage{makecell}
\usepackage{amsmath}
\usepackage{amssymb}
\usepackage{subcaption}
\usepackage{caption}
\usepackage{array}
\usepackage{tikz}

\def\bv{\mathbf{b}}

\def\dv{\mathbf{d}}
\def\hv{\mathbf{h}}

\def\KV{\mathbf{K}}
\def\qv{\mathbf{q}}

\DeclareMathOperator*{\softmax}{softmax}
\DeclareMathOperator{\AS}{AttnScore}

\title{Linear Recency Bias During Training Improves Transformers' Fit to Reading Times}

\author{Christian Clark \\ The Ohio State University \\ \texttt{clark.3664@osu.edu}
        \And  Byung-Doh Oh \\ New York University \\ \texttt{oh.b@nyu.edu}
        \And William Schuler \\ The Ohio State University \\ \texttt{schuler.77@osu.edu}}

\begin{document}
\maketitle
\begin{abstract}
Recent psycholinguistic research has compared human reading times to surprisal estimates from language models to study the factors shaping human sentence processing difficulty.
Previous studies have shown a strong fit between surprisal values from Transformers and reading times.
However, standard Transformers work with a lossless representation of the entire previous linguistic context, unlike models of human language processing that include memory decay.
To bridge this gap, this paper evaluates a modification of the Transformer model that uses ALiBi \citep{pressetal22}, a recency bias added to attention scores.
Surprisal estimates with ALiBi show an improved fit to human reading times compared to a standard Transformer baseline.
A subsequent analysis of attention heads suggests that ALiBi's mixture of slopes---which determine the rate of memory decay in each attention head---may play a role in the improvement by helping models with ALiBi to track different kinds of linguistic dependencies.
\end{abstract}

\definecolor{myblue}{rgb}{0.35,0.35,1}
\definecolor{myred}{rgb}{0.82,0.1,0.26}

\def\devardaHeatmap{%
{\setlength{\fboxsep}{0pt}%
\fbox{%
\begin{tikzpicture}[scale=0.8,baseline=(current bounding box)]
    \fontsize{7}{6}\selectfont

  \foreach \y [count=\n] in {
        {50/1,0/,0/},
        {25/$e^{-\lambda}$,50/1,0/},
        {13/$e^{-2\lambda}$,25/$e^{-\lambda}$,50/1},
    } {
      \foreach \x/\xtext [count=\m] in \y {
        \node[fill=myblue!\x!white, minimum size=8mm, text=black] at (\m,-\n) {\xtext};
      }
    }
\end{tikzpicture}%
}}}

\def\alibiHeatmap{%
{\setlength{\fboxsep}{0pt}%
\fbox{%
\begin{tikzpicture}[scale=0.8,baseline=(current bounding box)]
    \fontsize{7}{6}\selectfont
    \foreach \y [count=\n] in {
        {50/0,0/,0/},
        {25/$-1$,50/0,0/},
        {13/$-2$,25/$-1$,50/0},
    } {
      \foreach \x/\xtext [count=\m] in \y {
        \node[fill=myblue!\x!white, minimum size=8mm, text=black] at (\m,-\n) {\xtext};
      }
    }
\end{tikzpicture}%
}}}

\def\attnHeatmap{%
{\setlength{\fboxsep}{0pt}%
\fbox{%
\begin{tikzpicture}[scale=0.8,baseline=(current bounding box)]
    \fontsize{7}{6}\selectfont
    \foreach \y [count=\n] in {
        {36/$q_1k_1$,0/,0/},
        {18/$q_2k_1$,66/$q_2k_2$,0/},
        {44/$q_3k_1$,9/$q_3k_2$,57/$q_3k_3$},
    } {
      \foreach \x/\xtext [count=\m] in \y {
        \node[fill=myblue!\x, minimum size=8mm, text=black] at (\m,-\n) {\xtext};
      }
    }
\end{tikzpicture}%
}}}

\section{Introduction}

Expectation-based theories of human sentence processing \citep{hale01,levy08} posit that the difficulty of comprehending a word is proportional to its surprisal, i.e.\ negative log probability, given the preceding context.
This creates a natural interface between the task of language modeling, which estimates word probabilities, and modeling human sentence processing.
A range of studies \citep{goodkindbicknell18, wilcoxetal20, merkxfrank21} have compared surprisal estimates from families of language models (LMs) including n-gram models, recurrent neural networks, and Transformers, generally showing a strong fit between Transformer surprisal and psychometric data such as reading times.

\newlength{\oldtabcolsep}
\setlength{\oldtabcolsep}{\tabcolsep}

\setlength{\tabcolsep}{0pt}
\begin{figure}[t!]
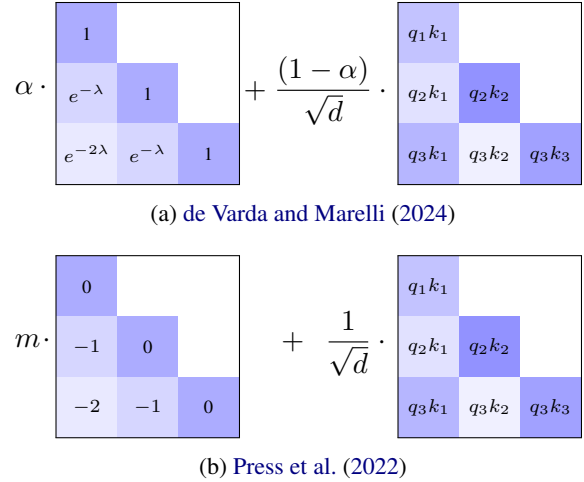

    \begin{subfigure}{\linewidth}
        \centering
         \begin{tabular}{ >{\centering}m{5mm} >{\centering}p{25mm} >{\centering}m{20mm} >{\centering}p{25mm} }
             $\alpha \cdot {}$ & \devardaHeatmap & \[+\ \frac{(1-\alpha)}{\sqrt{d}} \cdot {} \] & \attnHeatmap \\
        \end{tabular}    
        \vspace{-5mm}
        \caption{\protect{\citet{devardamarelli24}}}
        \vspace{3mm}
    \end{subfigure}
    \begin{subfigure}{\linewidth}
        \centering
         \begin{tabular}{ >{\centering}m{5mm} >{\centering}p{25mm} >{\centering}m{20mm} >{\centering}p{25mm} }
             $m \cdot {}$ & \alibiHeatmap & \[\ {} + \enskip \frac{1}{\sqrt{d}} \cdot {} \hspace{-3mm}\] & \attnHeatmap \\
        \end{tabular} 
        \vspace{-5mm}
        \caption{\protect{\citet{pressetal22}}}
    \end{subfigure}

    \caption{Illustration of two recency bias techniques tested in this work and defined in Equations~\eqref{eq:devarda} and \eqref{eq:alibi}. Bias matrices (left) are added to raw attention scores (right). Darker colors indicate higher scores. Hyperparameters $\alpha$, $\lambda$, and $m$ control the strength of the bias, and $\sqrt{d}$ scales the $q_ik_j$ values.}
    \label{fig:bias_comparison}
\end{figure}

The predictive power of surprisal estimates from Transformers raises the question of whether these models internally process language in a way that mirrors human language comprehension.
Indeed, the attention mechanism at the heart of the Transformer architecture bears a tantalizing similarity to models of comprehension based on cue-based retrieval \citep{ryulewis21,ohschuler22,timkeylinzen23}. 
However, the fact that a Transformer's context window---typically including hundreds or thousands of tokens---is fully retained in memory when predicting subsequent tokens seems unrealistic for modeling human memory; human working memory has a small capacity, and retrieval from longer-term memory is prone to decay and interference effects that Transformers do not explicitly model.\footnote{\citet{ryulewis21} do report facilitatory interference effects in GPT-2, wherein the presence of a distractor noun decreases surprisal at a target verb in ungrammatical sentences.~However, these effects seem unlikely to provide a general mechanism for humanlike forgetting in Transformers.}

We therefore consider altering Transformers' attention mechanism to include a \textit{recency bias} which upweights keys that are closer to a given query.
Such a bias brings Transformers more in line with cognitive models that include some notion of decay or lossy context \citep{baddeleyhitch74,baddeley03,lewisvasishth05,futrelletal20}.
The particular form of recency bias we test is based on ALiBi (Attention with Linear Biases), which was originally developed by \citet{pressetal22} as a method for helping Transformers to extrapolate beyond their context length.
Experiments on psycholinguistic corpora with reading times show that this modification to Transformer attention results in improved surprisal estimates compared to a standard Transformer, with ALiBi's mixture of slopes (decay rates specific to each attention head) playing an important role in the improvement.
A follow-up experiment suggests that the varied slopes may enable different attention heads in a model with ALiBi to track different linguistic dependencies.
Such results could have interesting implications for the implementation of memory decay in models of human language comprehension.\footnote{Code used in this paper's experiments will be made publicly available.}

\section{Related Work}
While neural LMs have been tested for some time as expectation-based models of human sentence processing \citep[e.g.][]{wilcoxetal20, ohetal22}, recent work has more specifically examined how these models' memory representations relate to processing difficulty.

One line of research draws a connection between the self-attention of Transformers \citep{vaswanietal17transformer} and cue-based retrieval models of sentence processing \citep[e.g.][]{lewisetal06}\footnote{This is because the dot product of representations (i.e.\ Eqn.~\ref{eq:attn}) is used in both models to quantify the degree of similarity \citep{merkxfrank21}.} and aims to derive measures from Transformer LMs that align with real-time processing behavior.
\citet{ryulewis21} define an attention entropy metric that quantifies the diffuseness of the attention weights over previous tokens, and shows patterns that are consistent with similarity-based interference observed during the processing of subject-verb agreement.
\citet{ohschuler22} propose a normalized attention entropy metric to control for the number of tokens in the previous context, as well as other predictors that capture the distance between attention weights of consecutive time steps, which are shown to be predictive of naturalistic reading times over a surprisal baseline.
\citet{timkeylinzen23} train an LM based on a modified version of the Simple Recurrent Network \citep{elman91} with one self-attention head that aggregates representations of previous words, which yields attention weights and surprisal that are sensitive to agreement and semantic attraction effects.

A second line of research is concerned with the interaction between memory-based effects and expectation-based effects, and broadly falls under the framework of lossy-context surprisal \citep{futrelletal20}.
Recent work has focused on ``corrupting'' the lossless representations of pretrained Transformer LMs and evaluating the quality of resulting surprisal estimates.
\citet{hahnetal22} implement a probabilistic erasure of words based on their frequency and position within the sentence, and show that the resulting surprisal estimates accurately predict the increased reading times at the main verb of deeply embedded sentences.
\citet{kuribayashietal22} constrain LMs' access to the previous context and report improvements in modeling naturalistic reading times of English and Japanese text.
\citet{devardamarelli24} incorporate a softer recency bias into the attention weights of Transformers for surprisal estimates that are more predictive of naturalistic reading times.

While there have been some promising results for modeling interference effects with neural LMs \citep{ryulewis21,ohschuler22}, this second line of work using simpler recency-based models may provide strong predictions for common cases that arise in broad-coverage modeling.
Additionally, as the vast majority of these results are based on post-hoc modifications to pretrained LMs like GPT-2 \citep{radfordetal19}, it remains to be seen how various constraints like recency biases influence LMs during training.
This work aims to address these gaps by newly training a set of LMs with various recency biases in a controlled setting, and evaluating their surprisal estimates across a wide range of naturalistic reading time corpora. 

\section{Background}

This section gives an overview of standard Transformer attention and how it is modified to incorporate recency biases in the experiments that follow.

\subsection{Transformer Attention Scores}

Within an autoregressive Transformer layer \citep{vaswanietal17transformer}, the attention sublayer calculates attention scores based on the scaled dot product between the $i$th query $\qv_i \in \mathbb{R}^d$ and the first $i$ keys $\KV \in \mathbb{R}^{d \times i}$:
\begin{equation} \label{eq:attn}
    \softmax \Big( \frac{\KV^\top\qv_i}{\sqrt{d}} \Big),
\end{equation}
where $i \in \{1, \dots, N\}$, $d$ is the dimension of the query and key, and $N$ is the sequence length.
The scaling factor $\sqrt{d}$ prevents the magnitude of the dot product from growing too large \citep{vaswanietal17transformer}.

\subsection{Recency Bias}

The two recency bias techniques tested in this paper---namely, the method from \citeauthor{devardamarelli24} (henceforth ``dVM bias'') and ALiBi---are illustrated in Figure~\ref{fig:bias_comparison}.
Both techniques involve modifications to the raw scores in Equation~\eqref{eq:attn}, before the softmax is taken.
The modifications have the effect of lowering attention scores to tokens more distant from the current query.

The dVM bias is implemented using a vector $\bv_i \in \mathbb{R}^i$ such that $\bv_i[j] = e^{-\lambda (i-j)}$ for $j \in \{1, \dots, i\}$.
The hyperparameter $\lambda$ determines the rate of decay.
A weighted combination is taken between the bias vector and the raw attention scores:
\begin{equation} \label{eq:devarda}
    \softmax \Big(\alpha \bv_i + (1-\alpha) \frac{\KV^\top \qv_i}{\sqrt{d}}\Big).
\end{equation}
The additional hyperparameter $\alpha$ determines the relative weight of the bias and original attention scores.

ALiBi uses a bias vector $\bv'_i \in \mathbb{R}^i$ such that $\bv'_i[j] = m \cdot (j-i)$ for $j \in \{1, \dots, i\}$.
The hyperparameter $m$ is a slope determining the rate of decay.
Slopes are defined separately for each attention head in a layer.
\citet{pressetal22} report optimal performance in input length extrapolation from setting the slope of head number $h$ out of $H$ total heads as
\begin{equation} \label{eq:slope}
m_j = 2^{-h \cdot 2^{(-\log_2 H + 3)}} =2^{-8h/H}.
\end{equation}
ALiBi bias is directly added to the raw attention scores:
\begin{equation} \label{eq:alibi}
    \softmax \Big(\bv'_i + \frac{\KV^\top \qv_i}{\sqrt{d}}\Big).
\end{equation}

Because the softmax operation involves exponentiation, the linear bias of ALiBi translates to an exponential decay in the final attention scores, consistent with the shape of the decay found in models like ACT-R \citep{lewisvasishth05}.
However, the terms in the dVM bias become doubly exponential after the softmax;\footnote{The softmax operation assigns the final attention score $\frac{e^{x_j}}{\sum_{k} e^{x_{k}}}$ to the $j$th key with raw score $x_j$. If the raw score is a weighted sum of the dVM bias term $e^{-\lambda(i-j)}$ and the scaled dot product between the query and key, then the final attention score will be proportional to $e^{e^{-\lambda(i-j)}}$.} we are not aware of any existing cognitive models that use this form of decay.

\section{Experiment 1: Recency Bias During Inference}

Experiment 1 considers the effect of incorporating a recency bias into an already trained Transformer LM.
Under this approach, the bias is included only at inference time, not during training.
This is the technique used by \citet{devardamarelli24}.

We evaluate surprisal estimates from a model with the dVM bias and a model with ALiBi.
These are compared against a baseline LM with no recency bias.

\subsection{Language Model} \label{sec:lm}

The design of the base language model used in this and subsequent experiments follows Pythia language models \citep{bidermanetal23}.
Pythia LMs are autoregressive, decoder-only Transformer models that vary primarily in their capacity and quantity of training data.
The main distinctions between Pythia LMs and other Transformer-based LM families are that Pythia LMs parallelize the computation of the self-attention sublayer and the feedforward neural network, and do not use shared parameters for the embedding and projection matrices.

The specific model configuration was chosen based on optimal settings found in previous work that compared reading time estimates from Pythia-style models with varying capacities and training data amounts \citep{ohschuler23emnlp}.
This model configuration uses two layers, four attention heads, and an embedding size of 256.
The training data amount was also determined based on the same work; it included the first 1,000 batches of the Pile \citep{gaoetal20}, a large collection of English-language datasets comprising approximately 300 billion tokens.
Each batch contains 1,024 examples with a sequence length of 2,048, for a total size of 2,097,152 training tokens across the first 1,000 batches.
While the model size and the amount of training data is much smaller than contemporary standards for Transformer LMs, this configuration was found to achieve strong fit to reading times \citep{ohschuler23emnlp} and has the additional benefit of allowing quick training of a variety of LMs.

Also following Pythia LMs, the base LM for this experiment uses rotary positional embeddings \citep{suetal24} and a context window of 2,048 tokens.
See Appendix \ref{sec:lm_training} for additional training details.

\subsection{Recency Biases} \label{sec:bias}

The LM with the dVM bias uses the hyperparameters $\lambda=82.86$ and $\alpha=0.37$, which were the optimal values found by \citet{devardamarelli24} in a grid search on the Provo corpus \citep{lukechristianson18} with GPT2-small as the base LM.\footnote{The present experiments used these hyperparameters for consistency with \citet{devardamarelli24}, but it is possible that other values of $\lambda$ and $\alpha$ could perform better on other corpora or LMs.}

The LM with ALiBi uses the slopes $1/4$, $1/16$, $1/64$, and $1/256$ for the four attention heads in each layer, following Equation~\eqref{eq:slope}.

\subsection{Corpora} \label{sec:corpora}
This experiment used reading times from six self-paced reading (SPR) and eye-tracking (ET) corpora, which are described below:
\begin{itemize}[leftmargin=*]
    \item Brown \citep{smithlevy13}: SPR times from 35 subjects that read 13 English passages from the Brown Corpus \citep{kucerafrancis67} consisting of a total of 7,188 words.
    \item Natural Stories \citep{futrelletal21}: SPR times from 181 subjects that read 10 naturalistic English stories consisting of a total of 10,256 words.
    \item UCL \citep{franketal13}: SPR times from 117 subjects and fixation durations from 48 subjects that read isolated sentences extracted from three novels written by aspiring authors, consisting of a total of 4,957 words.
    \item GECO \citep{copetal17}: Fixation durations from 14 monolingual subjects that read the English version of novel \textit{The Mysterious Affair at Styles} \citep{christie20} that consists of 13 chapters and 56,441 words.
    \item Dundee \citep{kennedyetal03}: Fixation durations from 10 subjects that read 67 English newspaper editorials consisting a total of 51,501 words.
    \item Provo \citep{lukechristianson18}: Fixation durations from 84 subjects that read 55 short English passages ranging between news articles, science magazines, and works of fiction consisting a total of 2,746 words.
\end{itemize}

For the SPR datasets, the by-word reading times were filtered to exclude those of sentence-initial and -final words and those shorter than 100 ms or longer than 3000 ms.
Additionally, the Natural Stories data from subjects who answered fewer than four comprehension questions correctly and the UCL SPR data from sentence-level trials with incorrect answers to comprehension questions  were removed.

\setlength{\tabcolsep}{3pt}
\begin{table}[t!]
    \centering
    \footnotesize
    \begin{tabular}{lrrr} \toprule
    Corpus/Measure & Fit & Exploratory & Held-out \\ \midrule
    Brown SPR & 59,292 & 29,671 & 30,157 \\
    Natural Stories SPR & 384,905 & 192,772 & 192,425 \\
    UCL SPR & 139,300 & 70,239 & 69,753 \\ \midrule
    UCL FP & 20,428 & 10,281 & 10,310 \\
    UCL GP & 20,428 & 10,281 & 10,310 \\
    GECO FP & 144,850 & 72,468 & 72,574 \\
    GECO GP & 144,850 & 72,468 & 72,574 \\
    Dundee SP & 155,483 & 77,809 & 77,101 \\
    Dundee FP & 98,115 & 48,598 & 48,794 \\
    Dundee GP & 98,115 & 48,598 & 48,794 \\
    Provo SP & 91,032 & 45,654 & 45,404 \\
    Provo FP & 52,959 & 26,539 & 26,640 \\
    Provo GP & 52,960 & 26,539 & 26,640 \\ \midrule
    Total & 1,462,717 & 731,917 & 731,476 \\ \bottomrule
    \end{tabular}
    \caption{Number of observations in each partition of each reading time corpus.}
    \label{tab:observations}
\end{table}
For the ET datasets, the by-word scan path (SP), first-pass (FP), and go-past (GP) durations were analyzed.\footnote{Refer to Appendix \ref{sec:et_measures} for their definitions. The SP duration could not be calculated for the GECO and UCL corpora that do not provide raw eye fixation durations.}
These datasets were filtered to remove data points for unfixated words, words following saccades longer than four words, and words at starts and ends of sentences and documents.
For the Dundee Corpus \citep{kennedyetal03} that further provides annotations of positions within lines and screens, data points corresponding to words at starts and ends of lines and screens were also excluded.

Prior to regression modeling, all datasets were split into fit, exploratory, and held-out partitions of roughly 50\%, 25\%, and 25\% of data points respectively.
This partitioning was conducted based on the sum of the subject index and the sentence index\footnote{If the sum of the subject and sentence number of a data point value is zero or one, modulo four, the data point was assigned to the fit partition; if this value is two, the data point was assigned to the exploratory partition; and if this value is three, the data point was assigned to the held-out partition.} in order to ensure that all data points from a subject reading a particular sentence are kept intact in a given partition.
The fit partition was used to fit the regression models, and all results are reported on the exploratory partition.
The held-out partition was reserved for statistical significance testing, and its use was kept to a minimum to minimize the need for multiple-trials correction.
The final number of observations in each partition of each corpus is summarized in Table \ref{tab:observations}.

\subsection{Linear Mixed-Effects Modeling} \label{sec:lmer}

This experiment fit a set of linear mixed-effects \citep[LME;][]{batesetal15} regression models to evaluate the influence of different recency biases on Transformer surprisal's fit to human reading times.
Following previous work \citep[e.g.][]{ohschuler23tacl, wilcoxetal23, shainetal24}, the increase in regression model log-likelihood ($\Delta$LogLik) due to including a surprisal predictor over a common baseline regression model was calculated on the exploratory partition of each dataset.

The baseline predictors are word length in characters, index of word position within each sentence, unigram surprisal (both SPR and ET corpora), as well as a whether the previous word was fixated (ET corpora only).
Unigram surprisal was estimated using the KenLM toolkit \citep{heafieldetal13} with default smoothing hyperparameters on the OpenWebText Corpus \citep{gokaslancohen19}, which contains about 6.5 billion whitespace-delimited words.
On top of these baseline regression models, surprisal at the current word and the previous word was included to capture lingering effects of the previous word \citep[i.e.~``spillover'' effects;][]{rayneretal83}.
Surprisal came from the LMs described in Sections \ref{sec:lm} and \ref{sec:bias}.
All regression models were fit to raw reading times; this assumption of a linear relationship between surprisal and reading times has recently received empirical support \citep{wilcoxetal23, xuetal23, shainetal24}.

The random effects structures of LME models were determined by starting with the maximal structure \citep{barretal13} and removing the least predictive effect iteratively until the models converged. 
The final random effects structure for LME models fit to SPR corpora included by-subject random slopes for word position, word length, and surprisal of current and previous word, a by-subject random intercept, and a by-sentence random intercept.
For ET corpora collected from a generally smaller number of subjects, the final random effects structure included random slopes for word position and surprisal of current word, a by-subject random intercept, and a by-sentence random intercept.

\begin{figure}[t]
    \centering
    \includegraphics[width=\linewidth]{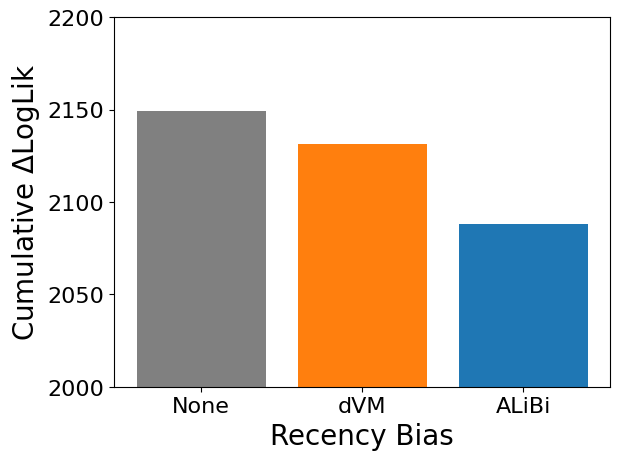}
    \caption{Aggregated likelihood results from Experiment 1. Improvements in log likelihood ($\Delta$LogLik) are summed across the corpora in Table~\ref{tab:observations}. Per-corpus results are in Appendix \ref{sec:per_corpus}.}
    \label{fig:exp1}
\end{figure}

\subsection{Results}

The results of Experiment 1 are presented in Figure~\ref{fig:exp1}, with $\Delta$LogLik summed over all corpora listed in Table \ref{tab:observations}.
On the whole, surprisal from LMs including a recency bias shows a worse fit to reading times than surprisal from the baseline LM with no recency bias, although the dVM bias performs somewhat better than ALiBi. 
We therefore do not generally replicate the improvements reported by \citet{devardamarelli24}, although improvements over the baseline LM are seen in the Dundee FP, Dundee GP, UCL SPR, UCL FP, and UCL GP corpora (see Appendix \ref{sec:per_corpus} for per-corpus results).

\section{Experiment 2: Recency Bias During Training and Inference}

Only introducing a recency bias during inference, as is done in Experiment 1, may be disadvantageous because it creates a mismatch between training and inference.
In addition, the original experiments testing ALiBi on input length extrapolation \citep{pressetal22} include this bias during both training and inference.
Experiment 2 therefore compares the LMs tested in the previous experiment with a similar set of LMs that include recency bias during both training and inference.

\subsection{Procedures}

Two new LMs were trained for this experiment: one including the dVM bias during training and one including ALiBi during training.
Following \citet{pressetal22}, rotary positional embeddings were removed from the LMs trained with recency bias.
Otherwise, LM training followed the setup described in Section \ref{sec:lm}, and LME models were fit using the corpora and procedures outlined in Sections \ref{sec:corpora} and \ref{sec:lmer}.
The newly trained models were compared against the dVM and ALiBi models with inference-only recency bias from Experiment 1.

\begin{figure}[t]
    \centering
    \includegraphics[width=\linewidth]{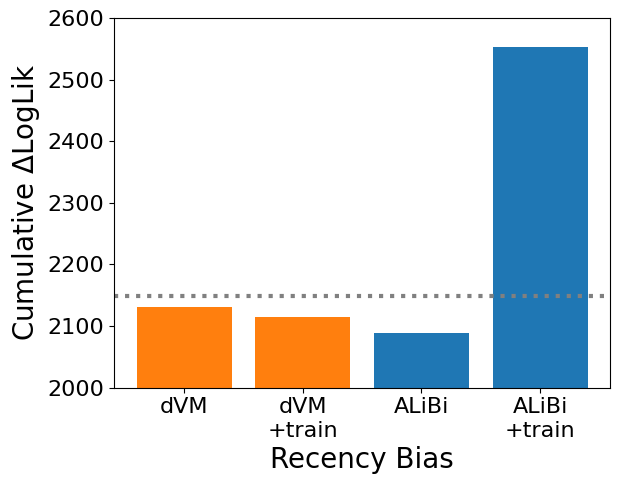}
    \caption{Aggregated likelihood results from Experiment 2. Models with a recency bias at inference time only (\textit{dVM}, \textit{ALiBi}) are compared against parallel models with a bias included during both training and inference (\textit{dVM+train}, \textit{ALiBi+train}). For comparison, the gray dashed line shows the cumulative $\Delta$LogLik from the baseline LM with no recency bias. Per-corpus results are in Appendix \ref{sec:per_corpus}.}
    \label{fig:exp2}
\end{figure}

\subsection{Results}

Figure~\ref{fig:exp2} presents the results, again aggregated over all corpora.
Including recency bias during training, instead of inference only, slightly decreases the $\Delta$LogLik from the dVM bias, but dramatically increases performance from ALiBi.
When ALiBi is included throughout training and inference, it attains a better $\Delta$LogLik than the baseline LM by a margin of over 400, which is significant at $p<0.001$ level by a permutation test of squared errors on the held-out partitions aggregated across all corpora.

\begin{figure}[t]
    \centering
    \includegraphics[width=\linewidth]{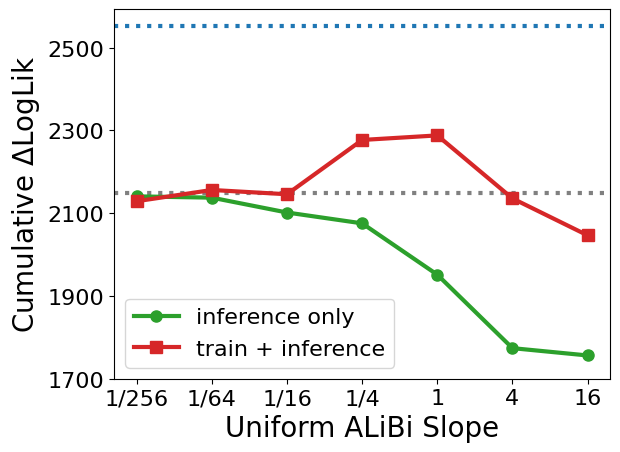}
    \caption{Aggregated likelihood results from Experiment 3. A variant of ALiBi in which all attention heads have the same slope was tested. One set of models included this bias at inference time only (green line), and the other set included the bias during both training and inference (red line). The gray dashed line shows the cumulative $\Delta$LogLik from a baseline LM with no recency bias, and the blue dashed line shows the same measure from an LM including ALiBi with mixed slopes during training and inference. Per-corpus results are in Appendix \ref{sec:per_corpus}.}
    \label{fig:exp3}
\end{figure}

\section{Experiment 3: Uniform ALiBi Slopes}

The version of ALiBi tested in the previous experiments uses different bias slopes for each attention head in a given layer, following Equation~\eqref{eq:slope}.
It might be asked whether this mixture of slopes is necessary for modeling human reading times, or if a single decay rate---as the dVM bias uses---can perform comparably well.
Experiment 3 tests this question by evaluating surprisal estimates from a set of LMs with a simplified version of ALiBi in which all attention heads use the same slope.

\subsection{Procedures}

Surprisal estimates were collected from a total of 14 LMs using ALiBi with uniform slopes, either during inference only as in Experiment 1 or during both inference and training.
The uniform slopes tested were $1/256$, $1/64$, $1/16$, $1/4$, $1$, $4$, and $16$.
For consistency with the models in Experiments 1 and 2, rotary embeddings were included in models with inference-only recency bias, but were removed from models that also included the recency bias during training. 
Other procedures followed those of Experiments 1 and 2.

\subsection{Results}

Results from this experiment are reported in Figure~\ref{fig:exp3}.
Two of the models with uniform ALiBi slopes ($m=1/4$ and $m=1$) during both training and inference performed better than the baseline LM with no recency bias.
However, none of the uniform-slope models matched the performance of the Experiment 2 model that included mixed-slope ALiBi during both training and inference (\textit{ALiBi+train} in Fig.~\ref{fig:exp2}).
These results indicate that including a mixture of slopes is necessary for obtaining optimal reading-time estimates from ALiBi.

\section{Experiment 4: Analysis of ALiBi Attention Heads}

The previous experiments show that including a mixture of head-specific slopes in ALiBi provides better reading time estimates than using a single slope across all attention heads.
Using mixture of slopes introduces a variable degree of recency bias across heads; we hypothesize that this is helpful for accessing relevant elements in the linguistic context that tend to appear at different distances from the current word.

\newcommand{\alb}[0]{\texttt{ALiBi-mix-TI}}

To test this hypothesis, we explored the sensitivity of the attention heads in the LM from Experiment 2 that includes mixed-slope ALiBi during both training and inference (henceforth \alb) to three types of semantic dependencies: first arguments (e.g.\ the relationship between a verb and its subject), second arguments (e.g.\ the relationship between a verb and its direct object), and coreference (e.g.\ the relationship between a pronoun and its antecedent).
We predicted that different attention heads would be sensitive to first and second argument dependencies (which tend to involve nearby words) and coreference dependencies (which often span longer distances).\footnote{A similar analysis of a model with uniform slope or no recency bias would be more arbitrary, since attention heads in such a model are not distinguished.} 

\subsection{Procedures}

The corpus used for this experiment was Natural Stories, which was selected because it had existing annotations with a generalized categorial grammar \citep{nguyenetal12,shainetal18:lincr} from which semantic dependencies could be extracted.
All instances of the three attachment operations---first argument, second argument, and coreference---in which the head of the dependency occurs later than the dependent were identified from the annotations.
Instances in which dependents occurred after heads were filtered out, since they are inaccessible in the masked attention used in autoregressive language models.
Across the 10 stories in the corpus, there were a total of 2,804 first-argument dependencies, 315 second-argument dependencies, and 1,428 coreference dependencies.
Each story entirely fit within the context window of \alb\ and therefore no dependencies crossed context windows.

Subsequently, for each dependency type, the mean attention score was calculated for each attention head in \alb.
This score came from averaging over the attention score from the query vector $\hv$ corresponding to the head of a dependency, to the key vector $\dv$ corresponding to the dependent:
\begin{equation}
    \frac{1}{\lvert D \rvert} \sum_{(\hv, \dv) \in D} \AS(\hv, \dv),
\end{equation}
where $D$ is the full set of dependencies of the relevant type and $\AS(\hv, \dv)$ is the attention score between the head and dependent according to the ALiBi formulation in Equation~\eqref{eq:alibi}.
In cases in which the head and/or dependent word spanned multiple tokens, an average was taken between tokens within words before the grand average was taken.

\begin{figure*}[ht!]
    \centering
    \begin{subfigure}{0.33\linewidth}
        \includegraphics[width=\linewidth]{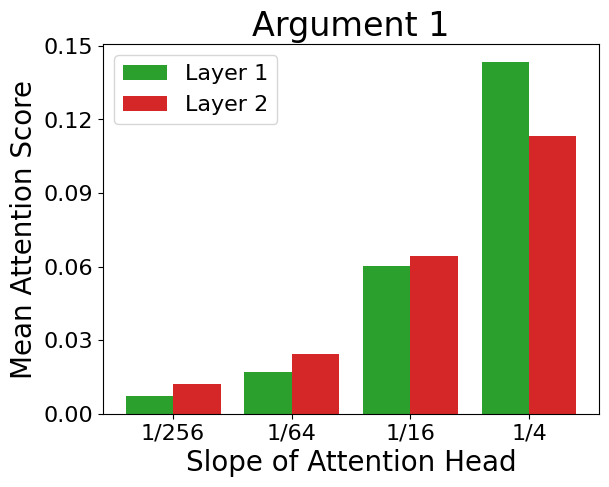}
        \label{fig:exp4_arg1}
    \end{subfigure}%
    \begin{subfigure}{0.33\linewidth}
        \includegraphics[width=\linewidth]{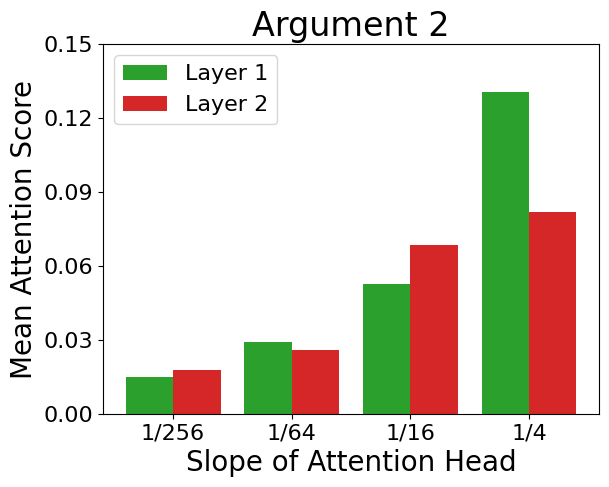}
        \label{fig:exp4_arg2}
    \end{subfigure}%
    \begin{subfigure}{0.33\linewidth}
        \includegraphics[width=\linewidth]{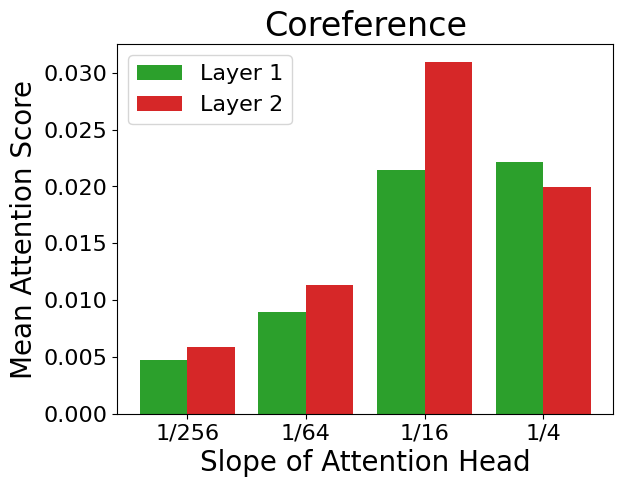}
        \label{fig:exp4_coref}
    \end{subfigure}
    \caption{Results from Experiment 4. Mean attention scores for three types of dependencies are presented for each attention head in a model with mixed ALiBi slopes. The evaluated model (\alb) includes two Transformer layers with four attention heads per layer.}
    \label{fig:exp4}
\end{figure*}

\subsection{Results}

Figure~\ref{fig:exp4} presents the mean attention scores for the three dependency relations across the eight attention heads in \alb.
For both first- and second-argument dependencies, attention heads with higher slopes (i.e.\ stronger recency bias) show higher attention scores than heads with lower slopes.
For coreference dependencies, however, the first Transformer layer shows similar attention scores from the heads with slopes $1/16$ and $1/4$, and the second layer has the highest mean attention score in the head with slope $1/16$.
This suggests that the model makes relatively greater use of an attention head with less decay for longer-distance coreference dependencies.
(The lower overall attention scores for coreference also reflect the longer average dependency distance; a larger number of intervening tokens compete for attention weight.)
The contrasting attention head behavior across argument and coreference dependencies supports the hypothesis that variable recency bias across heads is helpful for accessing different elements in the context.

\section{Discussion}

\begin{figure}[t!]
    \centering
    \includegraphics[width=\linewidth]{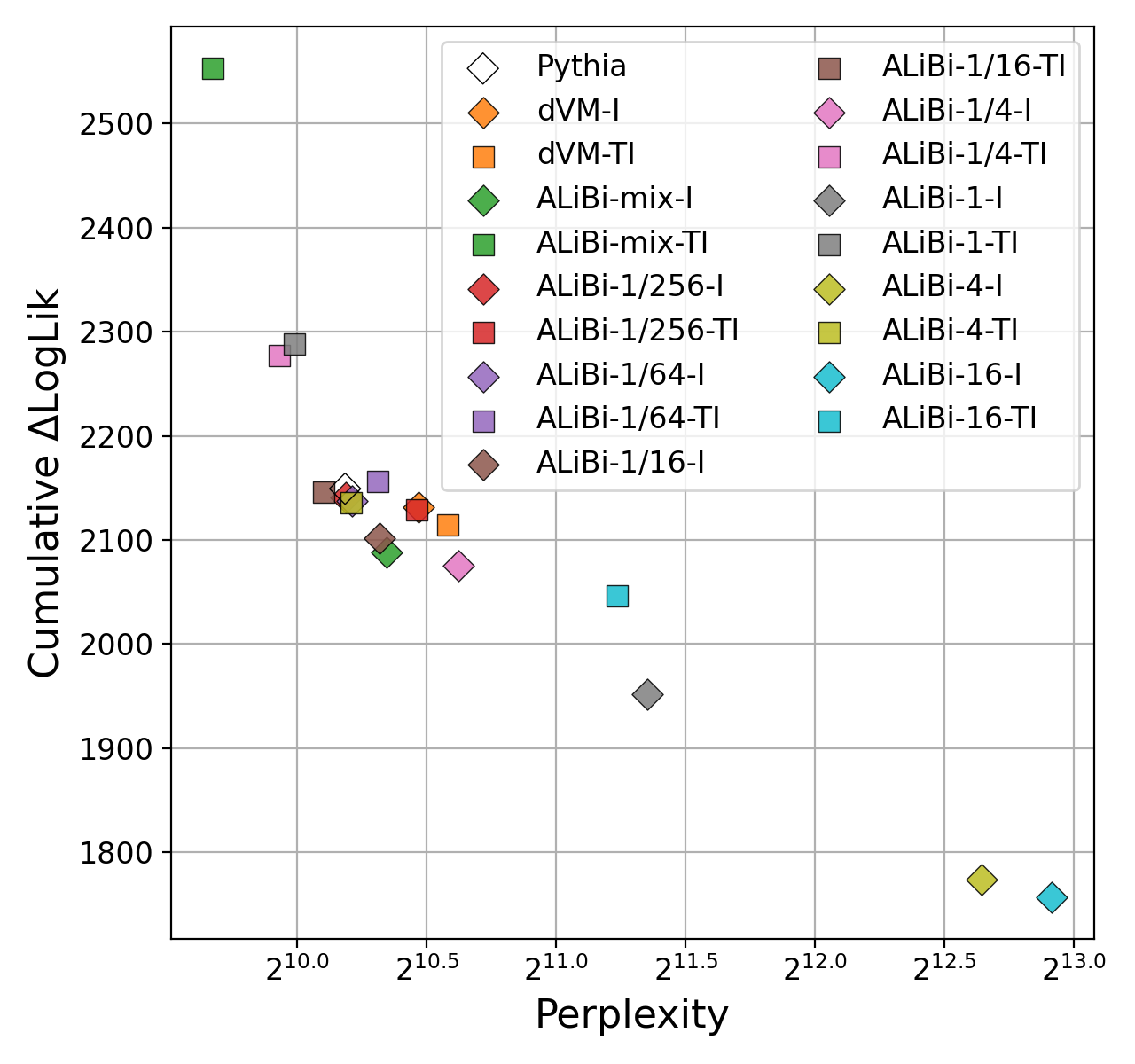}
    \caption{$\Delta$LogLik from LMs as a function of perplexity, both aggregated over all reading time corpora. In the legend, names ending in \textit{-TI} refer to LMs that include recency bias during training and inference, and names ending in \textit{-I} refer to models with recency bias during inference only.}
    \label{fig:ppl}
\end{figure}

Previous psycholinguistic work has reported conflicting findings about the relationship between the quality of an LM (measured in perplexity) and the psychometric predictive power of its surprisal estimates \cite{goodkindbicknell18, wilcoxetal23qp, ohschuler23tacl}.
Additionally, while \citet{pressetal22} show that ALiBi improves the perplexity of language models, \citet{devardamarelli24} report that their recency bias degrades language modeling performance.
Therefore, a natural question in the context of this work is how the experimental manipulations influence the perplexity of the LM.

Figure~\ref{fig:ppl} shows that \alb\ achieves lower perplexity than the baseline without recency bias, and that the other models tested in this work exhibit a negative relationship between perplexity and $\Delta$LogLik.
This suggests that the LMs examined in this work all lie in a regime where more accurate next-word predictions improve the fit of their surprisal estimates to human reading times \citep{ohschuler23emnlp}.
Moreover, given that the training setup such as the model capacity and training data is identical across conditions, this demonstrates the strong influence of recency biases on the probabilities learned by LMs during training, and opens new possibilities for modeling memory-based effects in sentence processing.

\section{Conclusion}

This work considers the effect of incorporating recency bias into a Transformer's attention mechanism, as a simple implementation of memory effects suitable for broad-coverage modeling.
Improvements in reading time prediction are observed from ALiBi, a biasing method originally developed for input length extrapolation.
Results are strongest when ALiBi is included during both training and inference, and when a mixture of slopes (memory decay parameters) is used across attention heads.
Analysis of individual attention heads provides evidence that different slopes may be helpful for tracking shorter- or longer-distance dependencies. 
The results suggest that incorporating varying rates of memory decay may be a promising direction both for improving language models and for developing cognitive models that can capture human behavior.

\section*{Limitations}
The alignment between surprisal estimates from Transformer language models and real-time comprehension difficulty presented in this work is based on language model variants trained on English text and data from subjects that are native speakers of English.
Therefore, the main findings of this work may not generalize to data collected in other languages.
Other possible limitations include the assumption of linear effects of surprisal in regression modeling.

\section*{Ethics Statement}
This work uses reading time data collected as part of previously published research \citep{smithlevy13, futrelletal21, franketal13, copetal17, kennedyetal03, lukechristianson18}.
Readers are referred to the respective publications for more information on the data collection and validation procedures.
As this work focuses on studying the connection between language models and mechanisms underlying human sentence processing, its potential negative impacts on society appear to be minimal.

\bibliography{custom}

\begin{thebibliography}{48}
\providecommand{\natexlab}[1]{#1}

\bibitem[{Andonian et~al.(2021)Andonian, Anthony, Biderman, Black, Gali, Gao, Hallahan, Levy-Kramer, Leahy, Nestler, Parker, Pieler, Purohit, Songz, Wang, and Weinbach}]{gptneox21}
Alex Andonian, Quentin Anthony, Stella Biderman, Sid Black, Preetham Gali, Leo Gao, Eric Hallahan, Josh Levy-Kramer, Connor Leahy, Lucas Nestler, Kip Parker, Michael Pieler, Shivanshu Purohit, Tri Songz, Phil Wang, and Samuel Weinbach. 2021.
\newblock \href {https://doi.org/10.5281/zenodo.5879544} {{GPT-NeoX}: Large scale autoregressive language modeling in {PyTorch}}.
\newblock \emph{Zenodo}.

\bibitem[{Baddeley(2003)}]{baddeley03}
Alan Baddeley. 2003.
\newblock \href {https://doi.org/10.1016/S0021-9924(03)00019-4} {Working memory and language: An overview}.
\newblock \emph{Journal of Communication Disorders}, 36(3):189--208.

\bibitem[{Baddeley and Hitch(1974)}]{baddeleyhitch74}
Alan~D. Baddeley and Graham Hitch. 1974.
\newblock \emph{{Working memory}}.
\newblock University of Stirling, Stirling, Scotland.

\bibitem[{Barr et~al.(2013)Barr, Levy, Scheepers, and Tily}]{barretal13}
Dale~J. Barr, Roger Levy, Christoph Scheepers, and Harry~J. Tily. 2013.
\newblock \href {https://doi.org/10.1016/j.jml.2012.11.001} {{Random effects structure for confirmatory hypothesis testing: Keep it maximal}}.
\newblock \emph{Journal of Memory and Language}, 68:255--278.

\bibitem[{Bates et~al.(2015)Bates, M{\"{a}}chler, Bolker, and Walker}]{batesetal15}
Douglas Bates, Martin M{\"{a}}chler, Ben Bolker, and Steve Walker. 2015.
\newblock \href {https://doi.org/10.18637/jss.v067.i01} {{Fitting linear mixed-effects models using lme4}}.
\newblock \emph{Journal of Statistical Software}, 67(1):1--48.

\bibitem[{Biderman et~al.(2023)Biderman, Schoelkopf, Anthony, Bradley, O'Brien, Hallahan, Khan, Purohit, Prashanth, Raff, Skowron, Sutawika, and van~der Wal}]{bidermanetal23}
Stella Biderman, Hailey Schoelkopf, Quentin~Gregory Anthony, Herbie Bradley, Kyle O'Brien, Eric Hallahan, Mohammad~Aflah Khan, Shivanshu Purohit, USVSN~Sai Prashanth, Edward Raff, Aviya Skowron, Lintang Sutawika, and Oskar van~der Wal. 2023.
\newblock \href {https://proceedings.mlr.press/v202/biderman23a.html} {Pythia: A suite for analyzing large language models across training and scaling}.
\newblock In \emph{Proceedings of the 40th International Conference on Machine Learning}, volume 202, pages 2397--2430.

\bibitem[{Christie(1920)}]{christie20}
Agatha Christie. 1920.
\newblock \href {https://www.gutenberg.org} {\emph{The mysterious affair at Styles}}.
\newblock John Lane.
\newblock Retrieved from Project Gutenberg.

\bibitem[{Cop et~al.(2017)Cop, Dirix, Drieghe, and Duyck}]{copetal17}
Uschi Cop, Nicolas Dirix, Denis Drieghe, and Wouter Duyck. 2017.
\newblock \href {https://doi.org/10.3758/s13428-016-0734-0} {{Presenting GECO: An eyetracking corpus of monolingual and bilingual sentence reading}}.
\newblock \emph{Behavior Research Methods}, 49(2):602--615.

\bibitem[{de~Varda and Marelli(2024)}]{devardamarelli24}
Andrea~Gregor de~Varda and Marco Marelli. 2024.
\newblock \href {https://aclanthology.org/2024.cmcl-1.3} {Locally biased transformers better align with human reading times}.
\newblock In \emph{Proceedings of the Workshop on Cognitive Modeling and Computational Linguistics}, pages 30--36.

\bibitem[{Elman(1991)}]{elman91}
Jeffrey~L. Elman. 1991.
\newblock \href {https://doi.org/10.1007/BF00114844} {{Distributed representations, simple recurrent networks, and grammatical structure}}.
\newblock \emph{Machine Learning}, 7:195--225.

\bibitem[{Frank et~al.(2013)Frank, Fernandez~Monsalve, Thompson, and Vigliocco}]{franketal13}
Stefan~L. Frank, Irene Fernandez~Monsalve, Robin~L. Thompson, and Gabriella Vigliocco. 2013.
\newblock \href {https://doi.org/10.3758/s13428-012-0313-y} {{Reading time data for evaluating broad-coverage models of English sentence processing}}.
\newblock \emph{Behavior Research Methods}, 45(4):1182--1190.

\bibitem[{Futrell et~al.(2020)Futrell, Gibson, and Levy}]{futrelletal20}
Richard Futrell, Edward Gibson, and Roger~P. Levy. 2020.
\newblock \href {https://doi.org/10.1111/cogs.12814} {Lossy-context surprisal: An information-theoretic model of memory effects in sentence processing}.
\newblock \emph{Cognitive Science}, 44(3).

\bibitem[{Futrell et~al.(2021)Futrell, Gibson, Tily, Blank, Vishnevetsky, Piantadosi, and Fedorenko}]{futrelletal21}
Richard Futrell, Edward Gibson, Harry~J. Tily, Idan Blank, Anastasia Vishnevetsky, Steven Piantadosi, and Evelina Fedorenko. 2021.
\newblock \href {https://doi.org/10.1007/s10579-020-09503-7} {{The Natural Stories corpus: A reading-time corpus of English texts containing rare syntactic constructions}}.
\newblock \emph{Language Resources and Evaluation}, 55:63--77.

\bibitem[{Gao et~al.(2020)Gao, Biderman, Black, Golding, Hoppe, Foster, Phang, He, Thite, Nabeshima, Presser, and Leahy}]{gaoetal20}
Leo Gao, Stella Biderman, Sid Black, Laurence Golding, Travis Hoppe, Charles Foster, Jason Phang, Horace He, Anish Thite, Noa Nabeshima, Shawn Presser, and Connor Leahy. 2020.
\newblock \href {https://arxiv.org/abs/2101.00027} {The {Pile}: An {800GB} dataset of diverse text for language modeling}.
\newblock \emph{arXiv preprint}, arXiv:2101.00027.

\bibitem[{Gokaslan and Cohen(2019)}]{gokaslancohen19}
Aaron Gokaslan and Vanya Cohen. 2019.
\newblock {OpenWebText Corpus}.
\newblock \url{http://Skylion007.github.io/OpenWebTextCorpus}.

\bibitem[{Goodkind and Bicknell(2018)}]{goodkindbicknell18}
Adam Goodkind and Klinton Bicknell. 2018.
\newblock \href {https://www.aclweb.org/anthology/W18-0102/} {{Predictive power of word surprisal for reading times is a linear function of language model quality}}.
\newblock In \emph{{Proceedings of the 8th Workshop on Cognitive Modeling and Computational Linguistics}}, pages 10--18.

\bibitem[{Hahn et~al.(2022)Hahn, Futrell, Gibson, and Levy}]{hahnetal22}
Michael Hahn, Richard Futrell, Edward Gibson, and Roger~P. Levy. 2022.
\newblock \href {https://doi.org/10.1073/pnas.2122602119} {A resource-rational model of human processing of recursive linguistic structure}.
\newblock \emph{Proceedings of the National Academy of Sciences}, 119(43):e2122602119.

\bibitem[{Hale(2001)}]{hale01}
John Hale. 2001.
\newblock \href {https://www.aclweb.org/anthology/N01-1021/} {{A probabilistic Earley parser as a psycholinguistic model}}.
\newblock In \emph{{Proceedings of the Second Meeting of the North American Chapter of the Association for Computational Linguistics}}.

\bibitem[{Heafield et~al.(2013)Heafield, Pouzyrevsky, Clark, and Koehn}]{heafieldetal13}
Kenneth Heafield, Ivan Pouzyrevsky, Jonathan~H. Clark, and Philipp Koehn. 2013.
\newblock \href {https://www.aclweb.org/anthology/P13-2121/} {{Scalable modified Kneser-Ney language model estimation}}.
\newblock In \emph{Proceedings of the 51st Annual Meeting of the Association for Computational Linguistics}, pages 690--696.

\bibitem[{Kennedy et~al.(2003)Kennedy, Hill, and Pynte}]{kennedyetal03}
Alan Kennedy, Robin Hill, and Joël Pynte. 2003.
\newblock {The Dundee Corpus}.
\newblock In \emph{Proceedings of the 12th European Conference on Eye Movement}.

\bibitem[{Kingma and Ba(2015)}]{kingmaba15}
Diederik~P. Kingma and Jimmy Ba. 2015.
\newblock \href {https://arxiv.org/abs/1412.6980} {{Adam: A method for stochastic optimization}}.
\newblock In \emph{Conference Track Proceedings of the 3rd International Conference on Learning Representations}.

\bibitem[{Kuribayashi et~al.(2022)Kuribayashi, Oseki, Brassard, and Inui}]{kuribayashietal22}
Tatsuki Kuribayashi, Yohei Oseki, Ana Brassard, and Kentaro Inui. 2022.
\newblock \href {https://aclanthology.org/2022.emnlp-main.712} {Context limitations make neural language models more human-like}.
\newblock In \emph{Proceedings of the 2022 Conference on Empirical Methods in Natural Language Processing}, pages 10421--10436.

\bibitem[{Ku\v{c}era and Francis(1967)}]{kucerafrancis67}
Henry Ku\v{c}era and W.~Nelson Francis. 1967.
\newblock \emph{Computational analysis of present-day American English}.
\newblock Brown University Press, Providence, RI.

\bibitem[{Levy(2008)}]{levy08}
Roger Levy. 2008.
\newblock \href {https://doi.org/10.1016/j.cognition.2007.05.006} {{Expectation-based syntactic comprehension}}.
\newblock \emph{Cognition}, 106(3):1126--1177.

\bibitem[{Lewis and Vasishth(2005)}]{lewisvasishth05}
Richard~L. Lewis and Shravan Vasishth. 2005.
\newblock \href {https://doi.org/10.1207/s15516709cog0000_25} {{An activation-based model of sentence processing as skilled memory retrieval}}.
\newblock \emph{Cognitive Science}, 29(3):375--419.

\bibitem[{Lewis et~al.(2006)Lewis, Vasishth, and Van~Dyke}]{lewisetal06}
Richard~L. Lewis, Shravan Vasishth, and Julie~A. Van~Dyke. 2006.
\newblock \href {https://doi.org/10.1016/j.tics.2006.08.007} {{Computational principles of working memory in sentence comprehension}}.
\newblock \emph{Trends in Cognitive Science}, 10(10):447--454.

\bibitem[{Luke and Christianson(2018)}]{lukechristianson18}
Steven~G. Luke and Kiel Christianson. 2018.
\newblock \href {https://doi.org/10.3758/s13428-017-0908-4} {{The Provo Corpus: A large eye-tracking corpus with predictability norms}}.
\newblock \emph{Behavior Research Methods}, 50(2):826--833.

\bibitem[{Merkx and Frank(2021)}]{merkxfrank21}
Danny Merkx and Stefan~L. Frank. 2021.
\newblock \href {https://doi.org/10.18653/v1/2021.cmcl-1.2} {Human sentence processing: Recurrence or attention?}
\newblock In \emph{Proceedings of the Workshop on Cognitive Modeling and Computational Linguistics}, pages 12--22.

\bibitem[{Nguyen et~al.(2012)Nguyen, van Schijndel, and Schuler}]{nguyenetal12}
Luan Nguyen, Marten van Schijndel, and William Schuler. 2012.
\newblock \href {https://www.aclweb.org/anthology/C12-1130/} {Accurate unbounded dependency recovery using generalized categorial grammars}.
\newblock In \emph{{Proceedings of the 24th International Conference on Computational Linguistics}}, pages 2125--2140.

\bibitem[{Oh et~al.(2022)Oh, Clark, and Schuler}]{ohetal22}
Byung-Doh Oh, Christian Clark, and William Schuler. 2022.
\newblock \href {https://doi.org/10.3389/frai.2022.777963} {Comparison of structural parsers and neural language models as surprisal estimators}.
\newblock \emph{Frontiers in Artificial Intelligence}, 5:777963.

\bibitem[{Oh and Schuler(2022)}]{ohschuler22}
Byung-Doh Oh and William Schuler. 2022.
\newblock \href {https://aclanthology.org/2022.emnlp-main.632} {Entropy- and distance-based predictors from {GPT-2} attention patterns predict reading times over and above {GPT-2} surprisal}.
\newblock In \emph{Proceedings of the 2022 Conference on Empirical Methods in Natural Language Processing}, pages 9324--9334.

\bibitem[{Oh and Schuler(2023{\natexlab{a}})}]{ohschuler23emnlp}
Byung-Doh Oh and William Schuler. 2023{\natexlab{a}}.
\newblock \href {https://aclanthology.org/2023.findings-emnlp.128/} {Transformer-based language model surprisal predicts human reading times best with about two billion training tokens}.
\newblock In \emph{Findings of the Association for Computational Linguistics: EMNLP 2023}, pages 1915--1921.

\bibitem[{Oh and Schuler(2023{\natexlab{b}})}]{ohschuler23tacl}
Byung-Doh Oh and William Schuler. 2023{\natexlab{b}}.
\newblock \href {https://doi.org/10.1162/tacl_a_00548} {Why does surprisal from larger {T}ransformer-based language models provide a poorer fit to human reading times?}
\newblock \emph{Transactions of the Association for Computational Linguistics}, 11:336--350.

\bibitem[{Press et~al.(2022)Press, Smith, and Lewis}]{pressetal22}
Ofir Press, Noah Smith, and Mike Lewis. 2022.
\newblock \href {https://openreview.net/forum?id=R8sQPpGCv0} {Train short, test long: Attention with linear biases enables input length extrapolation}.
\newblock In \emph{International Conference on Learning Representations}.

\bibitem[{Radford et~al.(2019)Radford, Wu, Child, Luan, Amodei, and Sutskever}]{radfordetal19}
Alec Radford, Jeff Wu, Rewon Child, David Luan, Dario Amodei, and Ilya Sutskever. 2019.
\newblock \href {https://cdn.openai.com/better-language-models/language_models_are_unsupervised_multitask_learners.pdf} {Language models are unsupervised multitask learners}.
\newblock \emph{OpenAI Technical Report}.

\bibitem[{Rajbhandari et~al.(2020)Rajbhandari, Rasley, Ruwase, and He}]{rajbhandarietal20}
Samyam Rajbhandari, Jeff Rasley, Olatunji Ruwase, and Yuxiong He. 2020.
\newblock \href {https://dl.acm.org/doi/10.5555/3433701.3433727} {{ZeRO}: Memory optimizations toward training trillion parameter models}.
\newblock In \emph{Proceedings of the International Conference for High Performance Computing, Networking, Storage and Analysis}, 20.

\bibitem[{Rayner et~al.(1983)Rayner, Carlson, and Frazier}]{rayneretal83}
Keith Rayner, Marcia Carlson, and Lyn Frazier. 1983.
\newblock \href {https://doi.org/10.1016/S0022-5371(83)90236-0} {{The interaction of syntax and semantics during sentence processing: Eye movements in the analysis of semantically biased sentences}}.
\newblock \emph{Journal of Verbal Learning and Verbal Behavior}, 22(3):358--374.

\bibitem[{Ryu and Lewis(2021)}]{ryulewis21}
Soo~Hyun Ryu and Richard~L. Lewis. 2021.
\newblock \href {https://aclanthology.org/2021.cmcl-1.6} {Accounting for agreement phenomena in sentence comprehension with {T}ransformer language models: Effects of similarity-based interference on surprisal and attention}.
\newblock In \emph{Proceedings of the Workshop on Cognitive Modeling and Computational Linguistics}, pages 61--71.

\bibitem[{Shain et~al.(2024)Shain, Meister, Pimentel, Cotterell, and Levy}]{shainetal24}
Cory Shain, Clara Meister, Tiago Pimentel, Ryan Cotterell, and Roger Levy. 2024.
\newblock \href {https://doi.org/10.1073/pnas.2307876121} {Large-scale evidence for logarithmic effects of word predictability on reading time}.
\newblock \emph{Proceedings of the National Academy of Sciences}, 121(10):e2307876121.

\bibitem[{Shain et~al.(2018)Shain, van Schijndel, and Schuler}]{shainetal18:lincr}
Cory Shain, Marten van Schijndel, and William Schuler. 2018.
\newblock \href {http://lrec-conf.org/workshops/lrec2018/W9/pdf/9_W9.pdf} {{Deep syntactic annotations for broad-coverage psycholinguistic modeling}}.
\newblock In \emph{Workshop on Linguistic and Neuro-Cognitive Resources}.

\bibitem[{Smith and Levy(2013)}]{smithlevy13}
Nathaniel~J. Smith and Roger Levy. 2013.
\newblock \href {https://doi.org/10.1016/j.cognition.2013.02.013} {{The effect of word predictability on reading time is logarithmic}}.
\newblock \emph{Cognition}, 128:302--319.

\bibitem[{Su et~al.(2024)Su, Ahmed, Lu, Pan, Bo, and Liu}]{suetal24}
Jianlin Su, Murtadha Ahmed, Yu~Lu, Shengfeng Pan, Wen Bo, and Yunfeng Liu. 2024.
\newblock \href {https://doi.org/10.1016/j.neucom.2023.127063} {Roformer: Enhanced transformer with rotary position embedding}.
\newblock \emph{Neurocomputing}, 568:127063.

\bibitem[{Timkey and Linzen(2023)}]{timkeylinzen23}
William Timkey and Tal Linzen. 2023.
\newblock \href {https://aclanthology.org/2023.findings-emnlp.582} {A language model with limited memory capacity captures interference in human sentence processing}.
\newblock In \emph{Findings of the Association for Computational Linguistics: EMNLP 2023}, pages 8705--8720.

\bibitem[{Vaswani et~al.(2017)Vaswani, Shazeer, Parmar, Uszkoreit, Jones, Gomez, Kaiser, and Polosukhin}]{vaswanietal17transformer}
Ashish Vaswani, Noam Shazeer, Niki Parmar, Jakob Uszkoreit, Llion Jones, Aidan~N. Gomez, {\L}ukasz Kaiser, and Illia Polosukhin. 2017.
\newblock \href {https://proceedings.neurips.cc/paper/2017/file/3f5ee243547dee91fbd053c1c4a845aa-Paper.pdf} {Attention is all you need}.
\newblock In \emph{Advances in Neural Information Processing Systems}, volume~30, pages 6000--6010.

\bibitem[{Wilcox et~al.(2020)Wilcox, Gauthier, Hu, Qian, and Levy}]{wilcoxetal20}
Ethan~Gotlieb Wilcox, Jon Gauthier, Jennifer Hu, Peng Qian, and Roger~P. Levy. 2020.
\newblock \href {https://cognitivesciencesociety.org/cogsci20/papers/0375} {On the predictive power of neural language models for human real-time comprehension behavior}.
\newblock In \emph{Proceedings of the 42nd Annual Meeting of the Cognitive Science Society}, pages 1707--1713.

\bibitem[{Wilcox et~al.(2023{\natexlab{a}})Wilcox, Meister, Cotterell, and Pimentel}]{wilcoxetal23qp}
Ethan~Gotlieb Wilcox, Clara Meister, Ryan Cotterell, and Tiago Pimentel. 2023{\natexlab{a}}.
\newblock \href {https://aclanthology.org/2023.emnlp-main.466} {Language model quality correlates with psychometric predictive power in multiple languages}.
\newblock In \emph{Proceedings of the 2023 Conference on Empirical Methods in Natural Language Processing}, pages 7503--7511.

\bibitem[{Wilcox et~al.(2023{\natexlab{b}})Wilcox, Pimentel, Meister, Cotterell, and Levy}]{wilcoxetal23}
Ethan~Gotlieb Wilcox, Tiago Pimentel, Clara Meister, Ryan Cotterell, and Roger~P. Levy. 2023{\natexlab{b}}.
\newblock \href {https://doi.org/10.1162/tacl_a_00612} {Testing the predictions of surprisal theory in 11 languages}.
\newblock \emph{Transactions of the Association for Computational Linguistics}, 11:1451--1470.

\bibitem[{Xu et~al.(2023)Xu, Chon, Liu, and Futrell}]{xuetal23}
Weijie Xu, Jason Chon, Tianran Liu, and Richard Futrell. 2023.
\newblock \href {https://aclanthology.org/2023.findings-emnlp.1052} {The linearity of the effect of surprisal on reading times across languages}.
\newblock In \emph{Findings of the Association for Computational Linguistics: EMNLP 2023}, pages 15711--15721.

\end{thebibliography}

\appendix

\section{Definition of Eye-Tracking Measures}
\label{sec:et_measures}
The following by-word eye-tracking measures were analyzed in this study:

\begin{itemize}[leftmargin=*]
    \item Scan path (SP) duration: Time taken after entering a word region from the left/right and before entering a different word region to the left/right.
    \item First-pass (FP) duration: Time taken after entering a word region from the left and before entering a different word region to the left/right.
    \item Go-past (GP) duration: Time taken after entering a word region from the left and before entering a word region to the right (including all regressive fixations).
\end{itemize}

\section{Training Procedures of Language Models}
\label{sec:lm_training}

All LMs used in the experiments were trained following the procedures of the Pythia LM variants using the GPT-NeoX library \citep{gptneox21}.
Training batches of 1,024 examples with a sequence length of 2,048 from the Pile \citep{gaoetal20} were provided to each model in the exact same order as the Pythia models.
The Zero Redundancy Optimizer \citep{rajbhandarietal20} implementation of Adam \citep{kingmaba15} with a learning rate of 0.001 was used to train the model parameters.
This learning rate was linearly warmed up over the first 1\% of training steps (i.e.\ 10 steps) and was annealed to a minimum of 0.0001 following a cosine schedule over the remainder of the 1,000 training steps.

\section{Per-Corpus LMER Results}
\label{sec:per_corpus}

$\Delta$LogLik values for each tested surprisal variant can be found in Table \ref{tab:per_corpus}.

\newcolumntype{R}{>{$}r<{$}}
\begin{table*}[ht!]
    \scriptsize
    \centering
    \begin{tabular}{l|RRRRRRRRRRRRR|R} 
    \toprule
        LM & \text{Brown} & \text{NS} & \text{UCL}_\text{SPR}  & \text{UCL}_\text{FP} & \text{UCL}_\text{GP} & \text{Geco}_\text{FP} & \text{Geco}_\text{GP} & \text{Dundee}_\text{SP} & \text{Dundee}_\text{FP} & \text{Dundee}_\text{GP} & \text{Provo}_\text{SP} & \text{Provo}_\text{FP} & \text{Provo}_\text{GP} & \text{Total} \\
                  \hline
        Pythia       & 499.0 & 66.0  & 57.1 & 38.6 & 19.5 & 311.6 & 138.7 & 144.2 & 508.1 & 272.9 & -27.1 & 101.3 & 19.4  & 2149.2 \\
        dVM-I       & 495.8 & 55.0  & 62.3 & 39.6 & 20.9 & 303.6 & 127.4 & 142.6 & 517.6 & 276.4 & -29.4 & 100.9 & 18.6  & 2131.4 \\
        dVM-TI  & 480.1 & 66.0  & 75.1 & 30.9 & 21.1 & 310.4 & 134.2 & 131.4 & 521.7 & 263.3 & -25.9 & 100.3 & 6.0   & 2114.6 \\
        ALiBi-mix-I           & 485.2 & 50.0  & 55.2 & 39.3 & 18.7 & 300.3 & 137.4 & 143.2 & 499.4 & 271.3 & -29.0 & 97.6  & 19.4  & 2088.0 \\
        ALiBi-mix-TI      & 519.5 & 109.0 & 68.1 & 54.7 & 36.8 & 472.0 & 228.1 & 144.8 & 507.2 & 274.8 & -30.9 & 136.5 & 32.8  & 2553.4 \\
        ALiBi-1/256-I        & 497.8 & 65.0  & 57.1 & 38.6 & 19.5 & 308.9 & 138.4 & 143.6 & 506.0 & 271.8 & -27.0 & 101.4 & 19.7  & 2140.8 \\
        ALiBi-1/256-TI   & 482.4 & 74.0  & 66.9 & 30.4 & 21.8 & 308.8 & 133.3 & 134.4 & 519.2 & 268.2 & -25.6 & 100.3 & 15.0  & 2129.0 \\
        ALiBi-1/64-I         & 495.4 & 60.0  & 68.7 & 38.7 & 19.4 & 306.3 & 138.7 & 142.2 & 502.7 & 270.4 & -27.1 & 101.6 & 20.6  & 2137.6 \\
        ALiBi-1/64-TI    & 492.3 & 72.0  & 96.9 & 39.8 & 19.6 & 302.4 & 134.5 & 134.6 & 500.2 & 256.0 & -22.4 & 110.2 & 20.1  & 2156.2 \\
        ALiBi-1/16-I         & 489.4 & 51.0  & 57.9 & 38.9 & 19.4 & 304.1 & 139.0 & 138.3 & 497.0 & 270.4 & -27.5 & 101.1 & 22.7  & 2101.7 \\
        ALiBi-1/16-TI    & 471.2 & 74.0  & 84.9 & 49.1 & 34.3 & 293.6 & 130.6 & 134.4 & 494.3 & 261.4 & -26.5 & 114.8 & 29.8  & 2145.8 \\
        ALiBi-1/4-I          & 472.9 & 40.0  & 85.1 & 39.5 & 19.6 & 313.1 & 144.1 & 133.1 & 476.4 & 263.8 & -29.5 & 96.5  & 20.9  & 2075.5 \\
        ALiBi-1/4-TI     & 477.1 & 87.0  & 84.6 & 53.1 & 29.9 & 334.6 & 160.2 & 139.6 & 491.0 & 272.0 & -24.1 & 132.4 & 39.6  & 2277.1 \\
        ALiBi-1-I          & 453.3 & 24.0  & 52.8 & 36.1 & 18.9 & 333.6 & 155.5 & 128.3 & 443.2 & 252.6 & -35.7 & 77.5  & 11.4  & 1951.4 \\
        ALiBi-1-TI     & 463.9 & 76.0  & 81.5 & 51.4 & 31.9 & 351.3 & 170.5 & 140.2 & 506.6 & 277.4 & -28.5 & 126.3 & 39.6  & 2288.1 \\
        ALiBi-4-I       & 442.4 & 11.0  & 46.5 & 18.8 & 7.2  & 335.0 & 153.5 & 104.2 & 423.0 & 234.3 & -39.4 & 48.6  & -11.3 & 1773.9 \\
        ALiBi-4-TI  & 464.1 & 76.0  & 95.0 & 48.4 & 33.0 & 294.1 & 130.8 & 139.0 & 487.6 & 263.6 & -31.8 & 114.6 & 21.4  & 2135.8 \\
        ALiBi-16-I      & 444.2 & 29.0  & 37.1 & 15.6 & 8.3  & 333.7 & 151.3 & 92.0  & 428.0 & 229.1 & -34.6 & 40.4  & -17.8 & 1756.2 \\
        ALiBi-16-TI & 463.8 & 49.0  & 96.3 & 23.1 & 16.0 & 326.8 & 139.1 & 115.9 & 499.9 & 262.6 & -29.4 & 94.0  & -10.6 & 2046.5 \\
    \bottomrule
    \end{tabular}
    \caption{$\Delta$LogLik of each surprisal variant tested in Experiments 1--3 on each corpus. \textit{NS} stands for Natural Stories. \textit{Pythia} is the baseline LM with no recency bias. LMs starting with \textit{dVM} use the bias technique from \citet{devardamarelli24}, and LMs starting with \textit{ALiBi} use the eponymous bias technique from \citet{pressetal22}. Models ending in \textit{-I} include recency bias at inference time only, while those ending in \textit{-TI} include recency bias during both training and inference. Models containing \textit{mix} use the mixture of attention head slopes recommended by \citet{pressetal22}, and models containing a number (1/256, 1/64, 1/16, 1/4, 1, or 4) use that number as a constant slope across all attention heads.}
    \label{tab:per_corpus}
\end{table*}

\end{document}